# Crack detection using tap-testing and machine learning techniques to prevent potential rockfall incidents

Roya Nasimi[*1], Fernando Moreu[2], John Stormont[3]

[1] Graduate Student, Department of Civil Construction and Environmental Engineering, University of New Mexico (UNM), Albuquerque, NM 87106
[2] Assistant Professor, Department of Civil Construction and Environmental Engineering, University of New Mexico (UNM), Albuquerque, NM 87106
[3] Professor, Department of Civil Construction and Environmental Engineering, University of New Mexico (UNM), Albuquerque, NM 87106

[*] Corresponding author, rhnasimi@unm.edu, 210 University of New Mexico, Albuquerque, NM 87131 0001.

## Abstract

Rockfalls are a hazard for the safety of infrastructure as well as people. Identifying loose rocks by inspection of slopes adjacent to roadways and other infrastructure and removing them in advance can be an effective way to prevent unexpected rockfall incidents. This paper proposes a system towards an automated inspection for potential rockfalls. A robot is used to repeatedly strike or tap on the rock surface. The sound from the tapping is collected by the robot and subsequently classified with the intent of identifying rocks that are broken and prone to fall. Principal Component Analysis (PCA) of the collected acoustic data is used to recognize patterns associated with rocks of various conditions, including intact as well as rock with different types and locations of cracks. The PCA classification was first demonstrated simulating sounds of different characteristics that were automatically trained and tested. Secondly, a laboratory test was conducted tapping rock specimens with three different levels of discontinuity in depth and shape. A real microphone mounted on the robot recorded the sound and the data were classified in three clusters within 2D space. A model was created using the training data to classify the reminder of the data (the test data). The performance of the method is evaluated with a confusion matrix.

## Introduction

Rockfalls are a daily hazard adjacent to roadways and rail lines in mountainous terrain. They are unpredictable in terms of magnitude and frequency which makes them dangerous [1,2]. Slope morphology, seismicity, human activities, and weathering rates are among the factors that triggers rockfalls. Depending on the slope gradient rockfall can occur in modes such as freefall, bouncing and rolling [3].

There are a wide variety of methods to predict potential rockfalls. Observational methods rely upon monitoring and documenting rockfall occurrences. Visual and imagery based rockfall and landslide assessment using photographs taken from a UAV or satellites is a common method [4]. Several researchers have made models and simulations combining 3D GIS models with rockfall programs to assess rockfall risks [5, 6]. Terrestrial Laser Scanning (TLS) has been used to estimate the location, scale, mechanism, and possible time of rockfall [7-16]. Moreover, LiDAR data have been used by researchers [17, 18] to interpret geological processes including rockfalls. However, these methods just obtain the before and after event information of the region, they are time consuming and require trained personnel [19].

Rock classification has been proposed to predict seismic rockfall [20]. Simple frictional models that employ digital terrain models (DTM) can be used to estimate rockfall prone locations as well as rock block movement [21-23]. Sophisticated 3-dimensional kinematic models can be used to predict potential rockfalls [24-27, 5] but require detailed inputs on rock and fracture properties that are often very difficult to obtain. Some researchers have used the Schmidt hammer tests data along with other test to characterize the rock material and rockfall trajectory in the field [28, 29] but these approaches require access to the rock surface.

A principal means to mitigate rock fall hazards is to detect and remove rocks that are prone to fall by manually inspecting and scaling existing exposed rock surfaces. Trained crews access the rock faces - often by rappelling over the edge from above or via portable lifts - and use pry-bars to strike the rock. The sound and feel of striking the rock are used to identify loose rocks that are then scaled [30, 31]. Rock inspection and scaling is high risk to the workers. It is costly in that requires a specialized crew, its time consuming and labour intensive. Further, it is necessary to close roads or divert traffic during inspection and scaling operations. A significant limitation of current inspection practice is that the results from striking the rock face is subjective as it is operator dependent. Further, the method is not conducive to recording data to allow for monitoring subtle changes in rock block response over time. Vogt et al. [30] developed an Electronic Sounding Device (ESD) that captures the sound from striking rock with a pry bar and processes it with a neural network model to distinguish the safe from an unsafe region. However, their method still requires access to the inspection surface to hit the rock.

To address some of the issues of manual rock face inspections, this research develops an automatic system for inspecting rock faces. An automated tap hammer mounted on a robot is used to strike a rock surface and record the resulting reflected waveforms with a microphone. The response of the rock to the hammer tap can be interpreted in terms of the stability of the rock, similar to the manual approach except that it is a less subjective measure. Sound data has been found to be conducive to assessing the condition of other materials (e.g., concrete) in conjunction with machine learning methods and convolutional neural network (CNN) models [32-34].

This paper presents the methodology to use sound data to classify rocks with respect to being unstable and prone to fall. The robot used to remotely collect the sound data is described. Laboratory tests were conducted on rock specimens with different discontinuity characteristics. The performance of the proposed method is evaluated by a confusion matrix.

## Methodology

### 1.1 Principal Component Analysis (PCA)

There are multiple methods to classify the sound data. This research suggests using Principal Component Analysis (PCA) method to identify the different type and depth of discontinuities in the rock surfaces. PCA of a data matrix transforms the space of the data vector and decrease the dimensionality of the multivariant data. This analysis extracts the dominant patterns and trends in the matrix and detects the outliers. PCA assumes that the data in new space is orthogonal to each other with zero covariance [37]. This method takes the data with multiple dependent variants and observations and represent the data in new space as orthogonal variables called principal components. PCA extracts the and keeps the most important information from the data table. Principal values are the calculated by linear combination of the original variables. The first principal value has the largest variation while the second has the second large variation and is orthogonal to the first principal value. Usually, the first two principal components are representative of the whole data because they have large inertia or called "explained". The inertia is the contribution of each principal value [35].

PCA analysis uses the singular value decomposition (SVD) concept to find the principal components and the transformed data. Assuming that the $A_{mxn}$ is the matrix of data, where m represents the number of observations and n represents the number of the variables (n<m), the data matrix can be written as Equation 1:

$$A_{m \times n} = V\Delta U^T \qquad \text{Equation 1}$$

Where, **V** and **U** are the left and right singular vectors and $\Delta$ is a diagonal matrix which components' is the square root of the Eigen values of $A \times A^T$. Matrix **U** gives the coefficients of the linear combinations, this matrix is calculated by combination of the eigen vectors. This matrix is a n × n square matrix where n is the number of the variables.

The scores matrix or the transformed data in the new space represented by $S$ is a $m \times n$ matrix that can be obtained using the Equations 2 and 3 as following.

$$S = V \times \Delta \qquad \text{Equation 2}$$

$$S = V \times \Delta \times U^T \times U = A \times U \qquad \text{Equation 3}$$

The **U** matrix is also called a projection matrix because with multiplying the data matrix with **U** data in the new space is calculated. The score matrix is a $m \times n$ matrix that shows each observation in new space. Usually, first two column of this matrix which corresponds to the first two principal components are the representative of all data and carries the most important information of the data.

The contribution of each principal component is calculated by a factor called explained or inertia. Considering that there are n principal components the explained variance of $k_{th}$ component can be calculated using Equation 4:

$$P = \frac{\sigma_k}{\sum_1^n \sigma} \qquad \text{Equation 4}$$

Where $\sigma$ represents the kth eigen value.

Figure 1 shows how the data obtained from each test can be organized to conduct the PCA. The data consists of tap sounds recorded using the microphone on the robot. In this data analysis, each tap is considered as observation and the number of points in the signal is considered as variation. The variation in the signal from the tap data is chosen to be n=100 points per tap experiment.

This research uses the data table from the experiments to classify the data from tapping different surfaces and cracks. Next, this research shows how the method can train a machine model that obtains taps from the surfaces and designate their classification without conducting a new PCA analysis on the testing data set. Finally, the method can also show data collected from new surfaces not included in training or testing data, capturing its relative location to known classified surfaces of the 2D principal component region.

| | Variable 1 | Variable 2 | ………. | Variable n |
|---|---|---|---|---|
| Tap1 | | | | |
| Tap 2 | | | | |
| ⋮ | | | | |
| | | | | |
| Tap m | | | | |

**Figure 1. Outline of data table for the analysis.**

This method divides the collected data into training and testing data and conduct a PCA analysis on the training data. The sound data are collected for a certain duration which include multiple hits data in it. Then the individual taps were extracted from the signal by processing it. processed the sound data and extracted individual tap signals from the whole signal, then set a threshold to extract the outliers and finally chose a certain number of the data points to consider for each tap. The PCA analysis is expected to classify the tap sounds using their signal variation in every tap. After creating the tap data table, a PCA was conducted using the selected training data. This paper adopted first two principal components to observe the data in 2-dimensional space. Then, used the kmeans clustering method to create trained regions and check if it is possible to know the testing data property and assign them to the right regions in trained areas. Following describes k-means clustering algorithm.

### 1.2 k-means clustering

There are multiple clustering algorithms to explore the structure of the data and categorize them into subgroups called clusters. Data clustering algorithms can be considered as unsupervised learning methods. One of the common algorithms to do data clustering is the k-means algorithm which is an iterative process [36]. This method creates regions in 2D PCA space of the data in which that each region belongs to a specific specimen type and the density of that specimen is higher in that region than the other regions. This method works with the minimization of the sum of the squares of the distances between the centroids of and the data points. Figure 2 shows the flowchart of the k-means algorithm. k-means clustering is an unsupervised learning method.

of the taps in the training data and the training data labels. Subsequently, considering that the c is the number of principal components representing the 95% of the variance of total data variance, the testing data scores were calculated using Equation 5.

$$S_t = (A_t - \mu) \times \mathbf{U}(:, \mathbf{1}:c) \qquad \text{Equation 5}$$

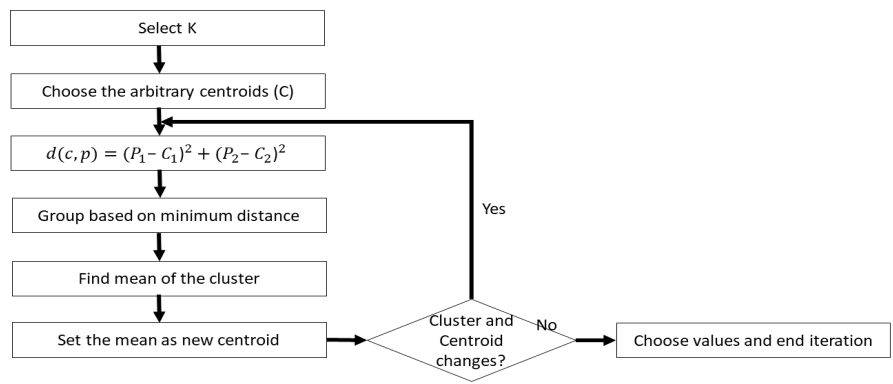

**Figure 2. K-means algorithm.**

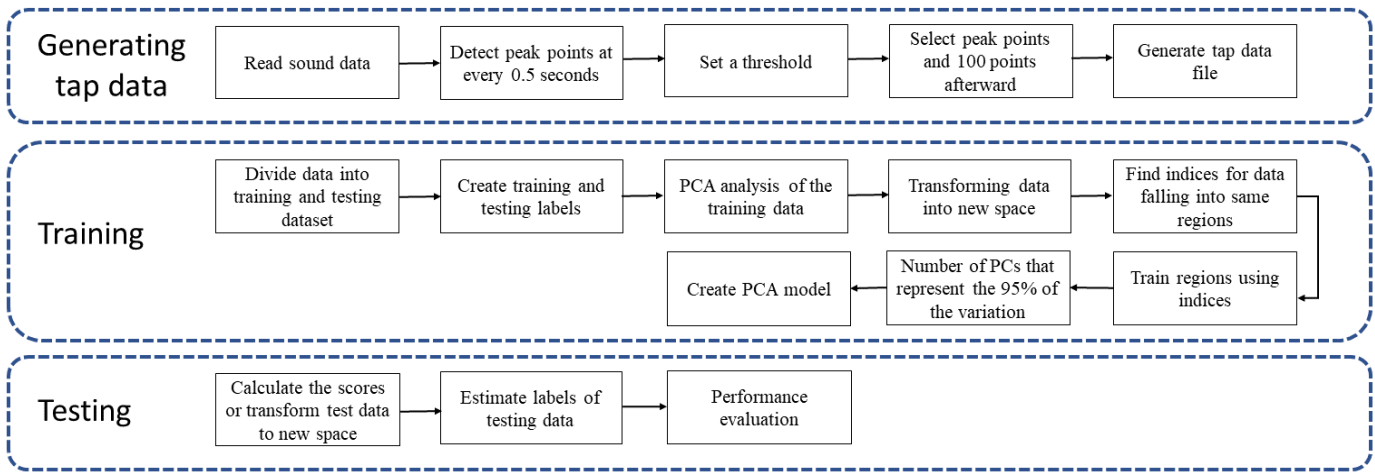

**Figure 3. Flowchart of the methodology.**

Finally, the testing data labels are defined using the calculated testing data score and the generated model. Figure 3 shows the flow chart of the proposed methodology.

After training the 2D PC space and assigning a respective area for each specimen, a model was created using the scores

Where, $\boldsymbol{S_t}$ is a $z \times 2$ matrix where z is the number of the observation or the taps in testing data, $\boldsymbol{A_t}$ is a testing matrix with z observation and n variables and $\mu$ is a vector with size of n representing the mean value of each tap signal. the mean of each column is subtracted from the respective column and is multiplied two first two PC of the training data.

The proposed method uses the confusion matrix to evaluate the performance of the estimation using the trained machine. Confusion or matching matrix lets us to visualize the performance in a table in which the horizontal and vertical directions correspond to predicted and true class values. The confusion matrix has $k^2$ components, where k is the number of the specimens or the expected clusters. The off-diagonal values of the matrix show the misclassified data while the diagonal components correspond to the taps or the observations that are correctly classified. In an ideal situation, when the error is zero, all off-diagonal values are zero in confusion matrix. The sum of the values showed in all cells is equal to the observation in testing dataset.

**Proof of concept simulation**

This research simulated five signals combining sine waves of different amplitudes and frequencies with a step signal with initial value of 1 dB and step time of 0.15 seconds. Each of five signals varied in amplitude and frequencies of the sine waves. For each signal 30 different sub-signals were generated with random noises, representing the taps. The five signals had the amplitude of frequencies of 1 dB and 20 rad/sec; 0.7 dB and 25 rad/sec; 0.4 dB and 15 rad/sec; 0.8 dB and 18 rad/Sec; and 1.5 dB and 13 rad/sec. PC analysis of data showed that, the first two PCs represent 66.5% and 26.7% of the whole data variation. Figure 4 (a) Figure 4(b) Figure 4(c) show the two PCs of each tap, training data and the trained regions using the training data, and the estimated testing data classification, respectively.

As illustrated in Figure 4 the method works perfectly in classification of the simulated signals in software environment. In the remainder of the paper, the method will be tested on signals collected from specimens and using the developed robot.

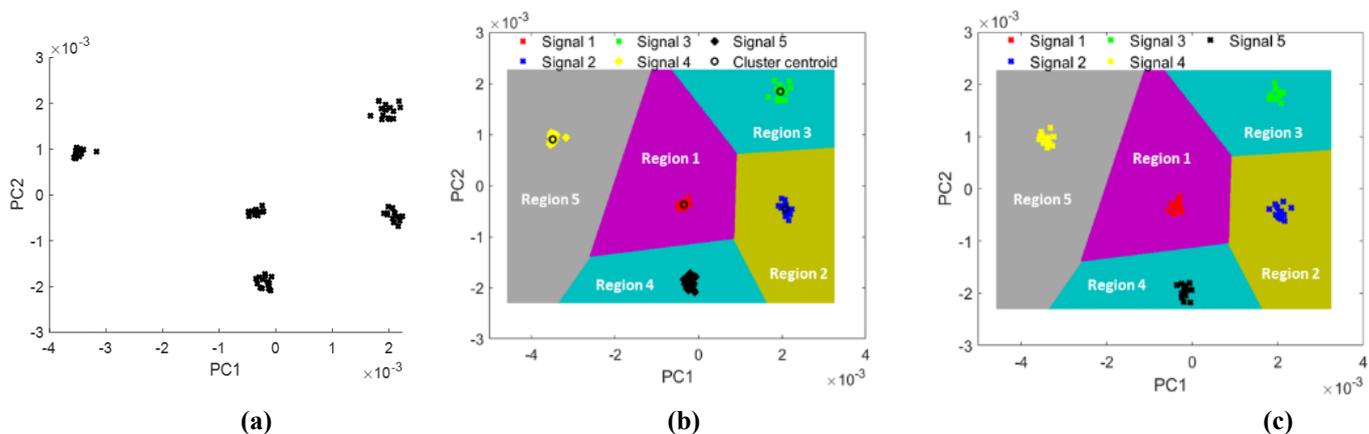

| (a) | (b) | (c) |
|---|---|---|

**Figure 4. Simulated signals: (a) two PC values of the signals; (b) trained 2D regions using training data; (b) projection of estimated testing data on trained regions.**

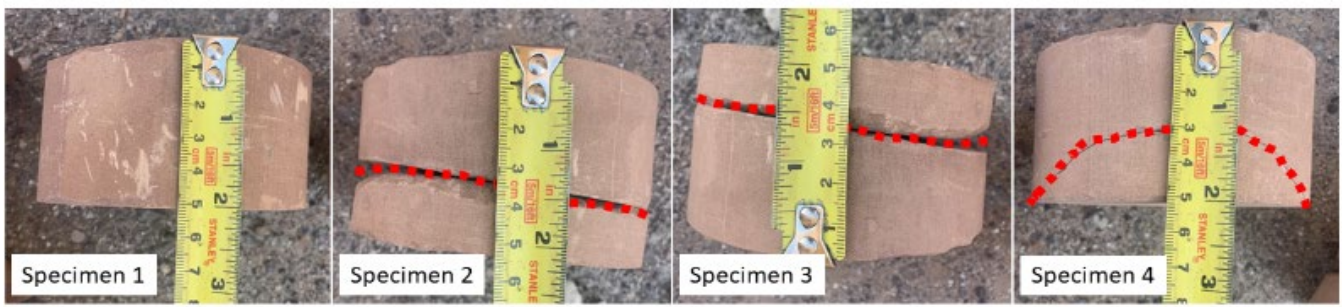

**Figure 5. Cylindrical rock specimen with four different crack configurations. View is from above.**

**Laboratory experiments**

In order to have a proof of concept of the proposed methodology in the context of rockfalls, the research uses a representative, simplified case of different specimens with known properties. In this way, the specimens are representative of variations in cracks and conditions that can be interrogated by our method. In this case, the research aims to simulate the stability of the various specimens to a progressively increasing level of vibration. The level of vibration required for the sample to fall in the laboratory tests is assued to correlate approximately to the susceptibility for failure in the field. . By generating the same vibration for all specimens in controlled environments it is possible to evaluate whether PCA interrogation can detect unstable specimens.

### 3.1 Specimen's description

Cylindrical specimens from the regional stone quarry were used. The specimens had a diameter and thickness of 10 cm and 5 cm, respectively. The team created cracks on the specimens. There were three specimens with four different Specimens as shown in Figure 5. Specimen 1 had no crack on it while specimen 2 had a continuous crack at an approximate depth of 2 cm from the surface (the front of cylinder) while specimen 3 was identical to the specimen 3 but in the reverse direction to consider for different crack depth the specimen 4 had a small partial discontinuity on the surface.

### 3.2 Shake table experiment

The response of the rock specimens under a dynamic load using a shake table was monitored to provide an indication of the stability of the different rock specimens under loads which may correspond to seismic loading. Two signals on the shake

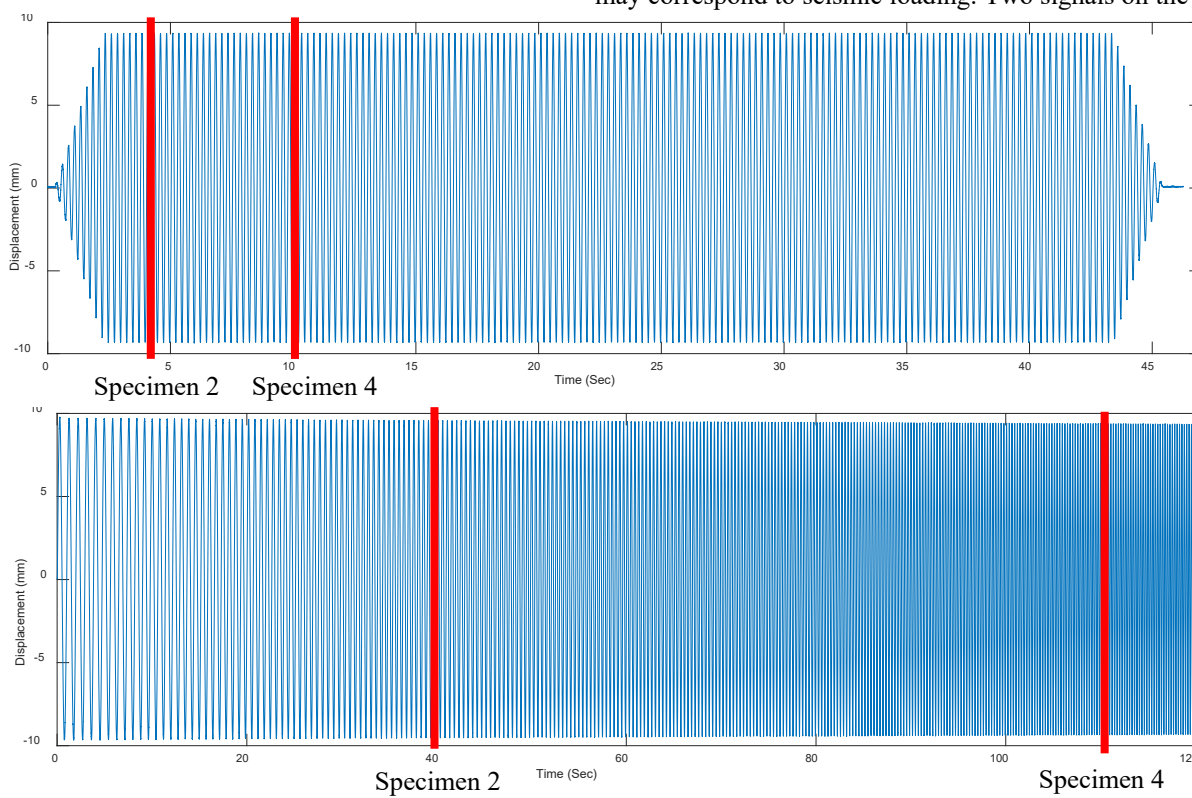

**Figure 6. Input signals to the shake table and the specimens' failure points; (above) Sinusoidal with frequency of 4 Hz; (below) Sine sweep with frequency of 1 Hz - 4 Hz**.

table were actuated when the rock specimens were placed on the table and recorded the falling order of the specimens. Figure 6 shows the input signals to the shake table. The first signal was a sinusoidal signal with an amplitude of 10 mm and frequency of 4 Hz and the second signal was a sine sweep with a constant amplitude of 10 mm but starting and ending frequency of 1 Hz and 4 Hz, respectively. Figure 7(a) and Figure 7(b) show the tests conducted with the shake table with the rock falling orders when the rock specimens were stimulated under sinusoidal, and sine sweep signals.

In these experiments specimens are placed with order of Specimen 1, Specimen 2, and Specimen 4 from left to right, respectively. The red lines in Figure 6 mark the approximate time that each rock failed and specimens' falling orders are labelled the in Figure 7.

The outcome of this preliminary experiment helped to understand the behaviour of the selected specimens under shaking vibration. In both tests specimen 2 fell first, specimen 4 fell after specimen 2 and specimen 1 didn't experience fall during the loading.

### 3.3 Remote tapping system

A remote-controlled robot was developed to automate the data collection. This robot uses the concept of tap testing for concrete mounted on a robot car. A remotely controlled tapping hammer system with four-bar crank rocker mechanism is designed to replace the manual tapping procedure. This robot called Brutus consists of the tapping hammer and data acquisition system mounted on a remotely controlled vehicle. The machine approaches to the surface and the hammer hits the surface of interest with a certain force and creates acoustic response. The tapping hammer is a 19 mm diameter steel ball. This hammer is connected to the end of a rocker arm. The robot could hit the surface with a specific speed, energy during the experiments which decreased the human factor error in the manual testing

For high quality data, a TASCAMDR-44 WL digital recorder was used to collect the sound data. The microphone collects the sound data in two channels over the frequency range of 100-24000 Hz. This robot was able to operate with 5V. Figure 8 shows the picture of the robot and its components that was developed for this research.

### 3.4 Tap testing laboratory experiments

To test these specimens, rock specimens were placed on a fixed base and placed it at a height that the impact hammer can hit. Operator drove the car closer to the specimens and hit the surface while recording the sound data in an SD card. The team conducted 12 tests in total, 3 tests on each surface each test lasted about 40 seconds hitting the surface approximately 70-75 times. Figure 9 shows an image from the experiment conducted at laboratory.

### 3.5 Sandstone specimen classification with different crack conditions

The sound data were collected with the sampling rate of 44100 Hz. Brutus tapped the surface at an approximate interval of 0.6 second. The team cut each sound data to get the time frame that the impact hammer starts and ends tapping on the surface. The peak points of the data were detected at each 0.5 second time interval. The method sets a threshold for the amplitude to remove the abnormal peaks that are smaller or larger than the normal tapping peak due to bad hits and selected 100 points of the signal after the peak point to consider for the variation pattern of the taps in that frame.

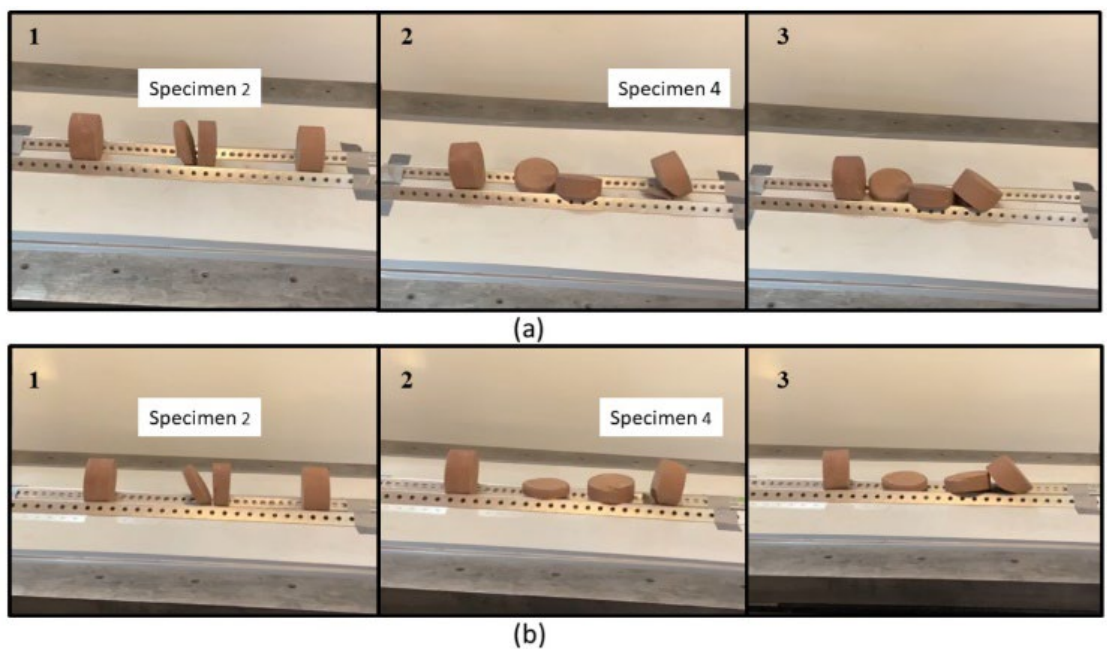

**Figure 7. Shake table experiment; (a) Rockfall sequence for the sinusoidal signal; (b) Rockfall order for sine sweep signal.**

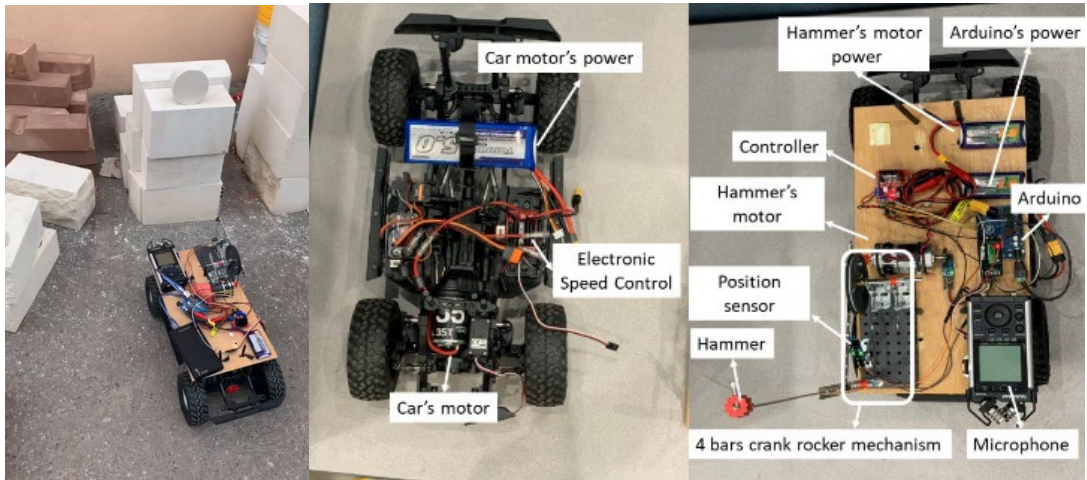

**Figure 8. Brutus I; (a) Brutus conducting an experiment (a) Base truck chassis of Brutus; (b) upper view of the sensors used on Brutus.**

3.5.1 Classification of two surfaces

For evaluating the algorithm in recognition of the intact and cracked surfaces the researchers selected tests from Specimen 1 and Specimen 2, respectively. The research team selected 60 percent of the data from each surface for the training and the remaining of the data for testing. PC analysis of training data taps calculated the two principal components of training data, Figure 10. Following table shows the number of the taps selected for each surface during the training and testing.

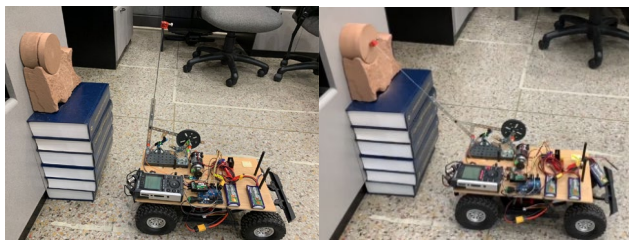

**(a)**                    **(b)**

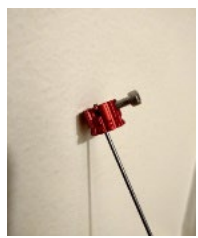

**(c)**

**Figure 9. Test conducted using Brutus: (a) pre-tap position; (b) tap event; (c) tip detail.**

**Table 1. Information of the training data.**

| Selected experiments | Training taps | Testing taps | Correctly classified |
|---|---|---|---|
| Specimen 1 | 44 | 29 | 29 |
| Specimen 2 | 43 | 29 | 29 |
| Total | 87 | 58 | 58 |

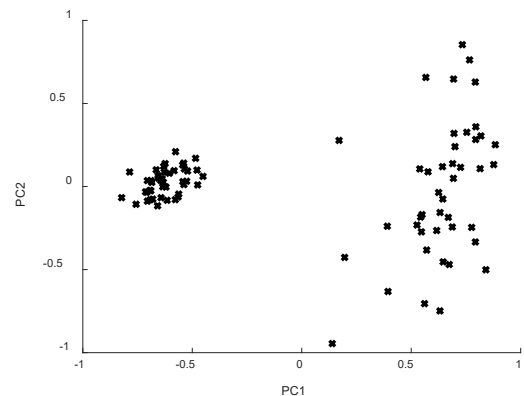

**Figure 10. First principal values of the training data.**

Subsequently, we found the center of clusters, classified the training data, and trained the regions using k-means method. The method was then applied to the testing data to check the quality of the classification (Figure 11). It worth mentioning that the tap sounds of the sound collected from the intact specimens are denser while the data from the cracked specimen is sparser.

The confusion matrix shown in Figure 12 shows the classification performance of the PCA analysis for classifying two different specimens' data collected using Brutus. As seen in Figure 11 (b) there's only one tap that fell past the border between the two regions, however the points are labelled correctly by the model and the border lines of the regions are approximate. For this data combination first two principal components represented the 41.6% and 8.8% of the variation of the whole data variation. During several experiments using the microphone or recording the data using a mobile, it was determined that the quality of the data effects the performance of the classification algorithm, therefore it is suggested to use a high-quality microphone.

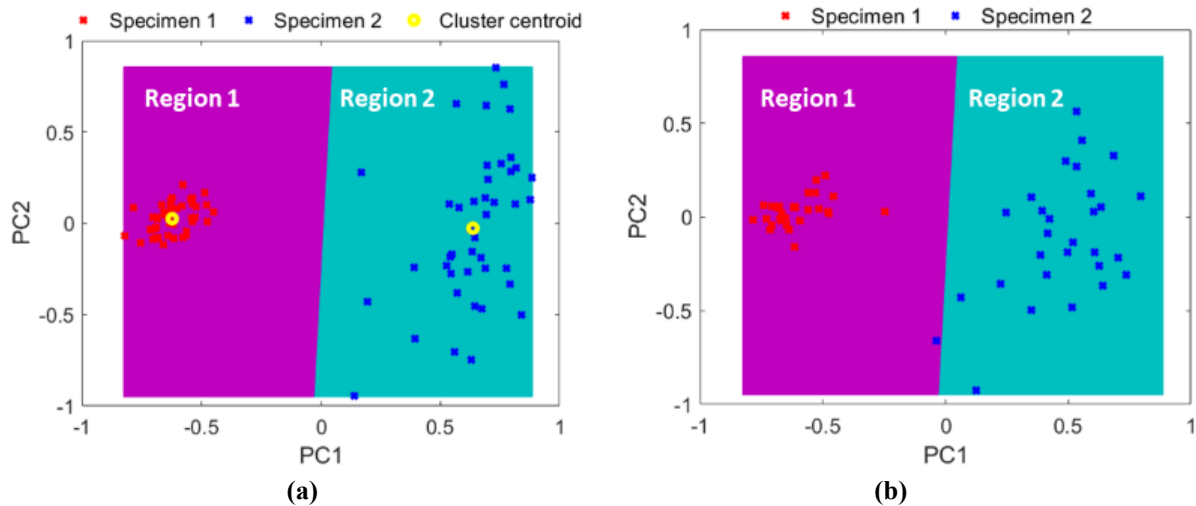

**Figure 11.** Data classification; (a) training data; (b) Testing data.

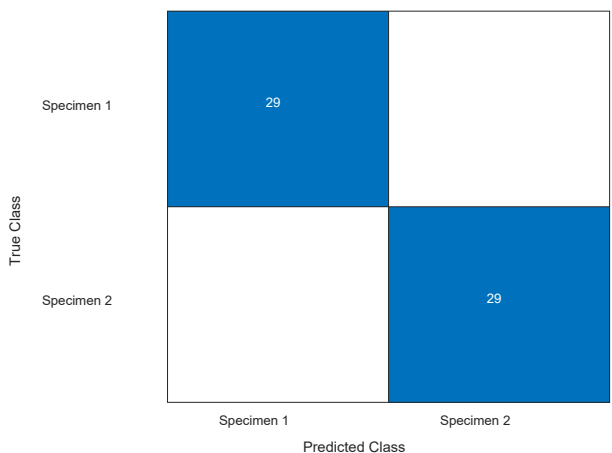

**Figure 12.** Confusion matrix of the testing data for two specimen classifications.

The confusion matrix shown in Figure 12 shows the classification performance of the PCA analysis for classifying two different specimens' data collected using Brutus. As seen in Figure 11 (b) there's only one tap that fell past the border between the two regions, however the points are labeled correctly by the model and the border lines of the regions are approximate. For this data combination first two principal components represented the 41.6% and 8.8% of the variation of the whole data variation. During several experiments using the microphone or recording the data using a mobile, it was determined that the quality of the data effects the performance of the classification algorithm, therefore it is suggested to use a high-quality microphone.

### 3.5.2 Classification of three surfaces

The first analysis demonstrated that the algorithm can distinguish between an intact and a cracked specimen. The next step was to determine if the algorithm can recognize the data from specimens with different crack characteristics or different depth. To this end, the authors selected three of the specimens to do the training and testing, Table 2 lists the detail of the taps used for analyzing three specimens. The first two PCs for this data set represented the 38.9% and 10.2% of the whole data variation, shown in Figure 13. The information of the tap data is listed in Table 2.

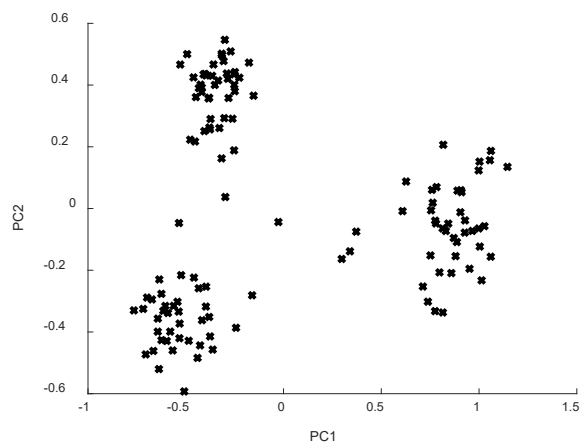

**Figure 13. Principal components of the training data.**

Figure 14 (a) shows the classification of training data into three clusters and the three regions designated to each specimen and Figure 14 (b) shows the projection of the estimated testing data on the trained regions.

The confusion matrix in Figure 15 shows the performance of the estimation as shown only two taps from specimen 3 are misclassified as specimen 1 and rest of the taps were correctly classified.

**Table 2. Information of the training and testing data of three specimens.**

| Selected experiments | Training taps | Testing taps | Correctly classified |
|---|---|---|---|
| Specimen 1 | 44 | 29 | 29 |
| Specimen 2 | 43 | 29 | 29 |
| Specimen 3 | 40 | 27 | 25 |
| Total | 127 | 85 | 83 |

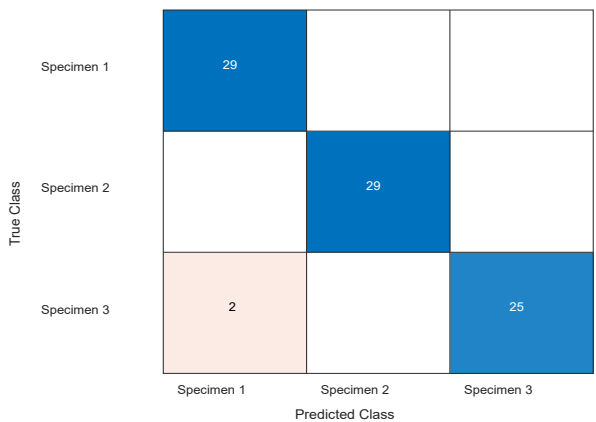

**Figure 15. Confusion matrix of the testing data for three specimen classifications.**

### 3.5.3 Condition assessment of an untrained new specimen

To check the performance of the method in classification of the surfaces that are new and were not included in training model, researchers tapped and collected data from Specimen 1 and Specimen 2 which were considered as most intact and most fractured specimens, respectively. The model was trained using the data from these two surfaces. Then the PCA values of the Specimen 4 were calculated and plotted in the 2D area. Figure 16 shows that when PC values of the Specimen 4 is calculated and plotted in the trained 2D space, that taps of this surface occupies a region in between the area of Specimen 1 and Specimen 2. Even though the method is not able to clearly classify a new rock specimen as broken or intact specimen, but it can still estimate the rock specimen's condition relative to other known specimens and this provides a qualitative estimate of the rock condition.

In summary, the proposed method and the system has demonstrated promising results in distinguishing different surfaces of material and different discontinuities from each other. Limitations of the method include : identifying the level of rock fall risk as it provides qualitative assessment of surface; and generating clear clusters when the expected clusters are more than three. For the future of the project, the team will analyze the field test data to: 1) identify problems and solutions related to field operations, 2) analyze the data from the rock surfaces that are prone to fall in the field to establish a relation between the PCA results and risk level in field, 3) improve the machine to classify wide spectrum of specimens using better quality and larger data set.

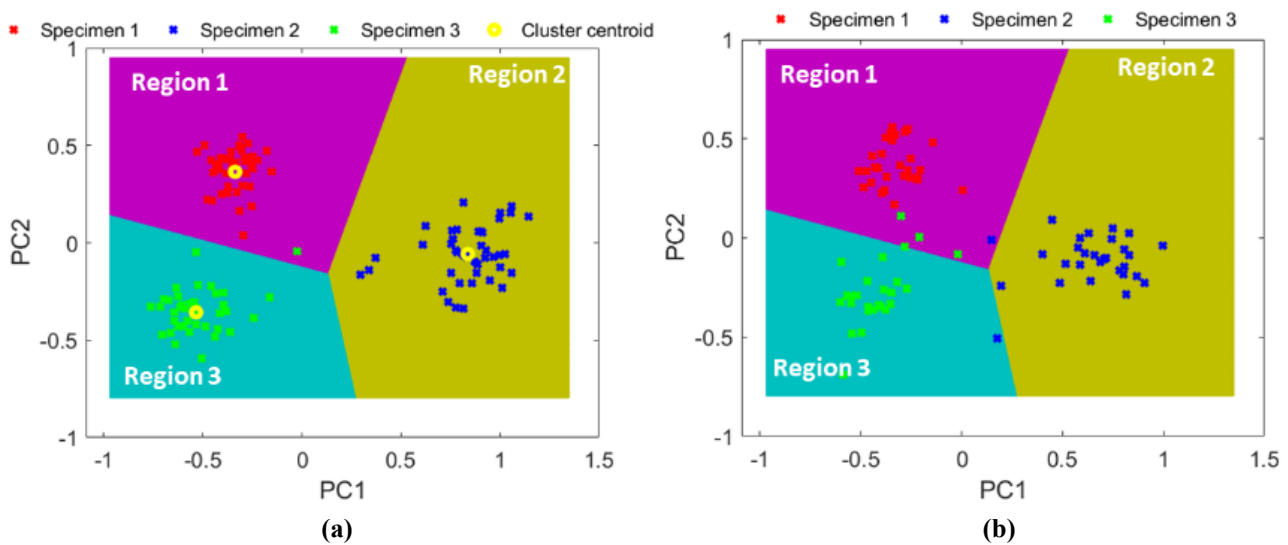

**Figure 14. Data classification; (a) training data; (b) Testing data.**

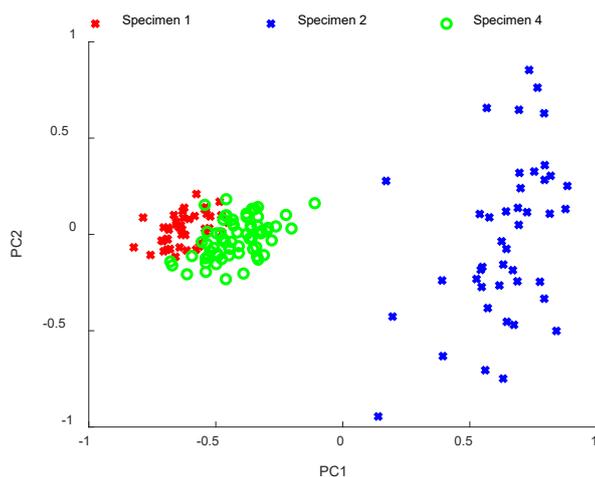

**Figure 16. Observation of an untrained data set (Specimen 4); Projection of the estimated untrained data values in trained region using training data (Specimen 1 and Specimen 2).**

## Conclusion

This research proposes an automated method that uses a remotely controlled robot system along with a machine learning algorithm to identify the rockfall risk of the rock zones. The method in this research automates the traditional tap testing procedure and mounts the hardware system on a robot and collect the acoustic data of the tapping hammer, then use the PCA analysis to classify the different rock specimens in different cluster. Subsequently, specimen clusters and their principal values are used to generate a machine. Later the testing data was classified in their respective clusters and was projected onto 2D principal component space without having a priori knowledge about the specimen number. Finally, a confusion matrix is used to evaluate the performance of the estimation using the testing data. This can classify rocks with different discontinuities and distinguished between an intact rock specimen from a rock specimen with shallow or deep crack in it. Moreover, the method was able to estimate the principal component values of tap sounds were not included in training process. Although the method was not able to label the untrained tap signal, it provided qualitative information about the untrained specimen compared to the specimens that were used to train the model. The promising result of this research demonstrates the potential of this approach to be a good alternative for manual rock inspections.